\documentclass{article}
 \pdfoutput=1

\usepackage[final,nonatbib]{neurips_2020}

\usepackage[utf8]{inputenc} 
\usepackage[T1]{fontenc}    
\usepackage{hyperref}       
\usepackage{url}            
\usepackage{booktabs}       
\usepackage{amsfonts}       
\usepackage{nicefrac}       
\usepackage{microtype}      
\usepackage{graphicx, caption}

\usepackage{amsmath,amsfonts,amssymb,amsthm}
\usepackage{mathtools}
\usepackage{commath}
\usepackage[sc,osf]{mathpazo}
\usepackage{booktabs,siunitx}
\usepackage{gensymb}
\usepackage{booktabs}
\usepackage{caption} 
\usepackage{footnote}
\usepackage{threeparttable}

\usepackage{authblk}
\usepackage{subfigure}
\usepackage{ifoddpage}
\usepackage{wrapfig}
\usepackage{comment}
\usepackage[numbers]{natbib}

\title{Using Convolutional Variational Autoencoders to Predict Post-Trauma Health Outcomes from Actigraphy Data}
\author[1,2]{\textbf{Ayse S. Cakmak}}
\author[2]{\textbf{Nina Thigpen}}
\author[2]{\textbf{Garrett Honke}}
\author[3]{\textbf{Erick Perez Alday}}
\author[3]{\textbf{Ali~Bahrami~Rad}}
\author[2]{\textbf{Rebecca Adaimi}}
\author[2]{\textbf{Chia Jung Chang}}
\author[3]{\textbf{Qiao Li}}
\author[2]{\textbf{Pramod Gupta}}
\author[4]{\textbf{Thomas Neylan}}
\author[5]{\textbf{Samuel A. McLean}}
\author[3,6]{\textbf{Gari D. Clifford,
on behalf of the AURORA investigators}}

\affil[1]{School of Electrical and Computer Engineering, Georgia Institute of Technology}
\affil[2]{Google X Development LLC}
\affil[3]{Department of Biomedical Informatics, School of Medicine, Emory University}
\affil[4]{Department of Psychiatry, University of California San Francisco, San Francisco, CA, USA}
\affil[5]{School of Medicine, University of North Carolina}
\affil[6]{Wallace H. Coulter Department of Biomedical Engineering, Georgia Institute of Technology and Emory University, Atlanta, GA, USA}

\begin{document}

\maketitle

\begin{abstract}
  Depression and post-traumatic stress disorder (PTSD) are psychiatric conditions commonly associated with experiencing a traumatic event. Estimating mental health status through non-invasive techniques such as activity-based algorithms can help to identify successful early interventions. In this work, we used locomotor activity captured from 1113 individuals who wore a research grade smartwatch post-trauma. A convolutional variational autoencoder (VAE) architecture was used for unsupervised feature extraction from four weeks of actigraphy data. By using  VAE  latent variables and the participant’s pre-trauma physical health status as features, a logistic regression classifier achieved an area under the receiver operating characteristic curve (AUC) of 0.64 to estimate mental health outcomes. The results indicate that the VAE model is a promising approach for actigraphy data analysis for mental health outcomes in long-term studies. 
\end{abstract}

\section{Introduction}
Mental health disorders are a significant problem in the United States and worldwide, with the number of adults struggling with mental illness increasing every day \cite{czeisler2020mental}. Two common mental health illnesses are Post-Traumatic Stress Disorder (PTSD) and depression, which affects 250 million people worldwide \cite{world2017depression}. PTSD symptoms include sleep disturbances, hyperarousal, and persistent intrusive memories of trauma, while depression symptoms include fatigue, markedly diminished interest in most activities, and depressed mood \cite{american2013diagnostic}. Wearable devices can detect alterations in daily activity and other behavioral patterns that could result from worsening symptoms.

Wearable devices are ideal tools for monitoring because they provide a passive data collection method that can track high-risk individuals and detect when a user's physical behavior becomes anomalous or suggests a negative change in prognosis. Actigraphy data has been used to estimate disturbances in sleep \cite{long2017actigraphy} and diagnostic, and severity information for mental health disorders \cite{tahmasian2013clinical, khawaja2014actigraphy, hvolby2008actigraphic}, though previous work suggests detecting sleep disturbances experienced by people with mental health disorders is markedly more difficult \cite{biddle2015accuracy, inman1990sleep, chung2010relationship}. Furthermore, actigraphy derived rest--activity metrics and variability have been previously investigated in depression cohorts \cite{burton2013activity, krafty2019measuring}. Deep learning methods could learn meaningful representations from the actigraphy data. Also, unsupervised learning is well suited for this problem space because the data without the labels derived from clinical surveys could be utilized. This research represents the first attempt to apply unsupervised deep learning methods to actigraphy data for feature extraction and representation.

In this work, we aim to develop models to estimate PTSD and depression as determined by clinical surveys using locomotor activity measured from a wearable device. We developed two models that used actigraphy data from a four-week window previous to the clinical surveys.

\section{Methods}
\subsection{Dataset and Preprocessing}
The data used in this work is part of the Advancing Understanding of RecOvery afteR traumA (AURORA) study dataset, which consists of individuals enrolled in the emergency department within 72 hours after experiencing a traumatic event \cite{mclean2020aurora}.
For the current study, we present the analysis of the data from 1113 participants enrolled between July 31, 2017, and July 31, 2019. Participants were $35\pm13.1$ years old and 36\% male. Traumatic events that qualified automatically for study enrollment were motor vehicle collision, physical assault, sexual assault, fall $>10$ feet, or mass casualty incidents. Participants were asked to wear a research wristwatch (Verily Life Sciences, San Francisco) at least 21 hours a day for the study period and at subsequent times that vary by the study participant.

\begin{figure}[b]
    \centering
    \includegraphics[width=0.80\textwidth]{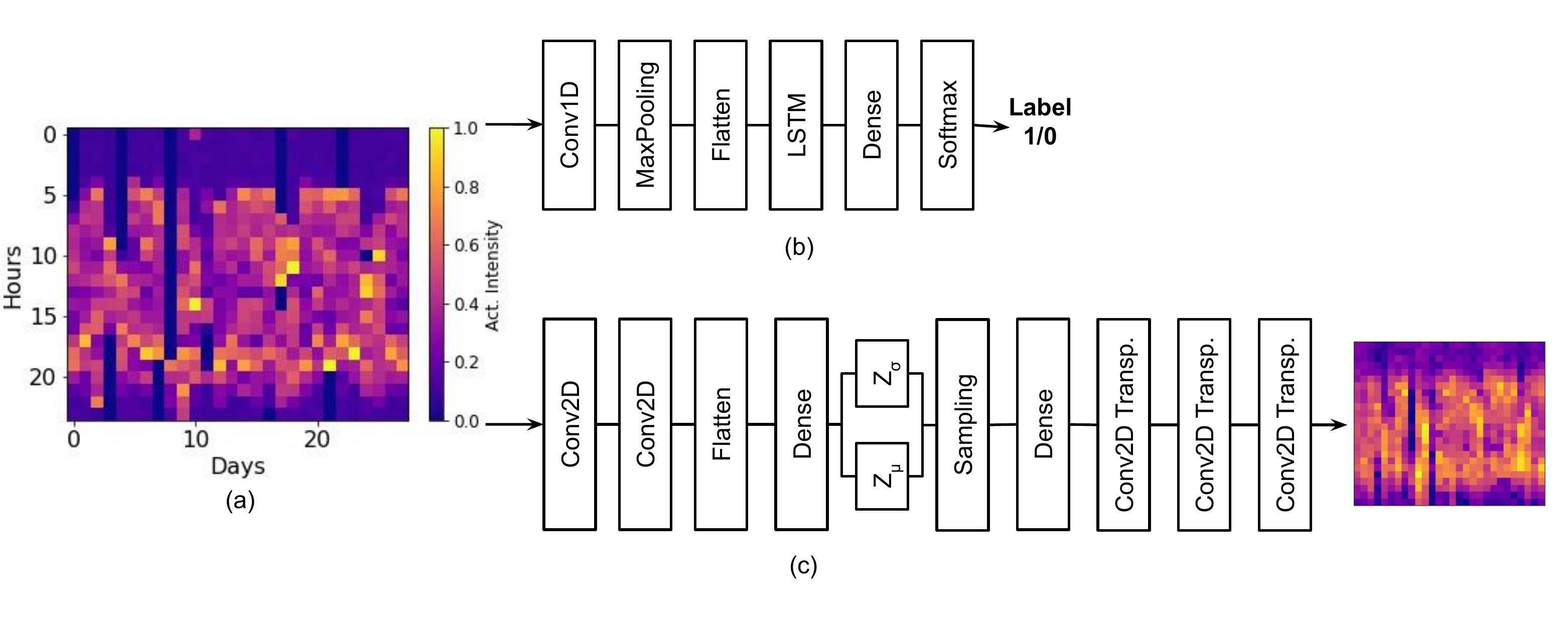}
    \caption{(a) Daily actigraphy levels for one participant. (b) CNN-LSTM and (c) variational autoencoder models.}
    \label{fused}
\end{figure}

Clinical follow-up surveys via web-based or phone assessments captured the mental health symptoms at eight weeks after initial evaluation. The Posttraumatic Stress Disorder Checklist for DSM-5 (PCL-5) was used to measure PTSD symptoms and participants with PCL-5 score greater than 28 were labeled as PTSD \cite{blevins2015posttraumatic}. Scored depression variables from PROMIS Depression - Short Form 8b (PROM-Dep8b) were used to measure depression symptoms, with a threshold of 60 for depression \cite{amtmann2014comparing}. One item from Pittsburgh Sleep Quality Index Addendum \cite{insana2013validation} (``how often did you awaken from sleep with severe anxiety or panic’’) was used to assess the difficulty in staying asleep (PanicSleep).  Also, the participant's general physical health status in the 30 days pre-trauma was used as a feature to test if it could increase the performance of models based on passive actigraphy data. This pre-health score is a derived normative score based on questions from the 12-Item Short Form Health Survey (SF-12) \cite{ware199612}.

The raw 3D accelerometer data were collected from the research wristwatch at a sampling frequency of 30 Hz.  These data were converted to activity counts which measure movement intensity. The Z-axis actigraphy data were bandpass filtered from 0.25-11 Hz to eliminate extremely slow or fast movements \cite{Ancoli-Israel2003}.  Then, the maximum values inside non-overlapping one-second windows were summed for each 30-second `epoch’ of data \cite{Borazio2014}. Fig. \ref{fused} (a) is a "double plot" that shows activity levels measured over 28 days. Each column of the plot is created by stacking two consecutive days of data. Participants with more than 50\% missingness were excluded from the analysis, and missing data sections were replaced with zeros.

\subsection{Models and Experimental Setup}
The outcomes of the clinical surveys were used to create the classes for the binary classification experiments. In the first experiment, the PCL-5 survey was used by itself, while in the second experiment, PCL-5 and PanicSleep survey scores were used. Lastly, all three surveys were combined to find the participants who experienced both depression and PTSD symptoms to create the unhealthy class. In the experiments, all participants from the unhealthy class were used, and the over-represented healthy class was under-sampled so that the results were not biased due to the unequal class prevalence. For the deep learning models, internal 5-fold cross-validation was performed. All experiments were repeated 30 times on external folds with different random samples from the majority (healthy) class.

 A variational autoencoder (VAE) is an unsupervised generative model that has an encoding phase in which the input data is projected onto lower-dimensional latent representations and a decoding phase that reconstructs the input, as shown in Fig. \ref{fused} (c). However, in the VAE model, the encoder is trained under the restriction that the latent representations follow a Gaussian distribution $N(Z_{\mu},Z_{\sigma})$. In this work, unlabelled actigraphy maps were used to train a convolutional VAE with two 2D convolution layers (Conv2D) with 16 and 32 number of filters and kernel sizes of 3. The number of units in the dense layers was set to 16. The number of filters in the Conv2DTranspose layers were 32, 16, and 1. The embedding dimension of VAE was 8. The model was trained for 30 epochs with a batch size of 128. Then, the latent representation of the actigraphy maps ($z_{act}$) was used as input features to a logistic regression model in binary classification experiments. 
 
 Secondly, an alternative supervised CNN-LSTM model was trained to estimate mental health outcomes from clinical surveys. The number of filters in the Conv1D layer was set to 32, and the kernel size was 3. The number of units in the LSTM and the dense layer was set to 20. Actigraphy data were inputted as 24-hour subsequences, and the model was trained for 30 epochs with a batch size of 32. 
 
 Lastly, 100 healthy and 100 unhealthy artificial actigraphy maps were generated with VAE models by using randomly sampled encoding vectors. The artificial data was used in the training step of the CNN-LSTM model to test if the performance will be improved.

\section{Results and Discussion}

\begin{figure}[htp]
    \centering
    \includegraphics[width=0.55\textwidth]{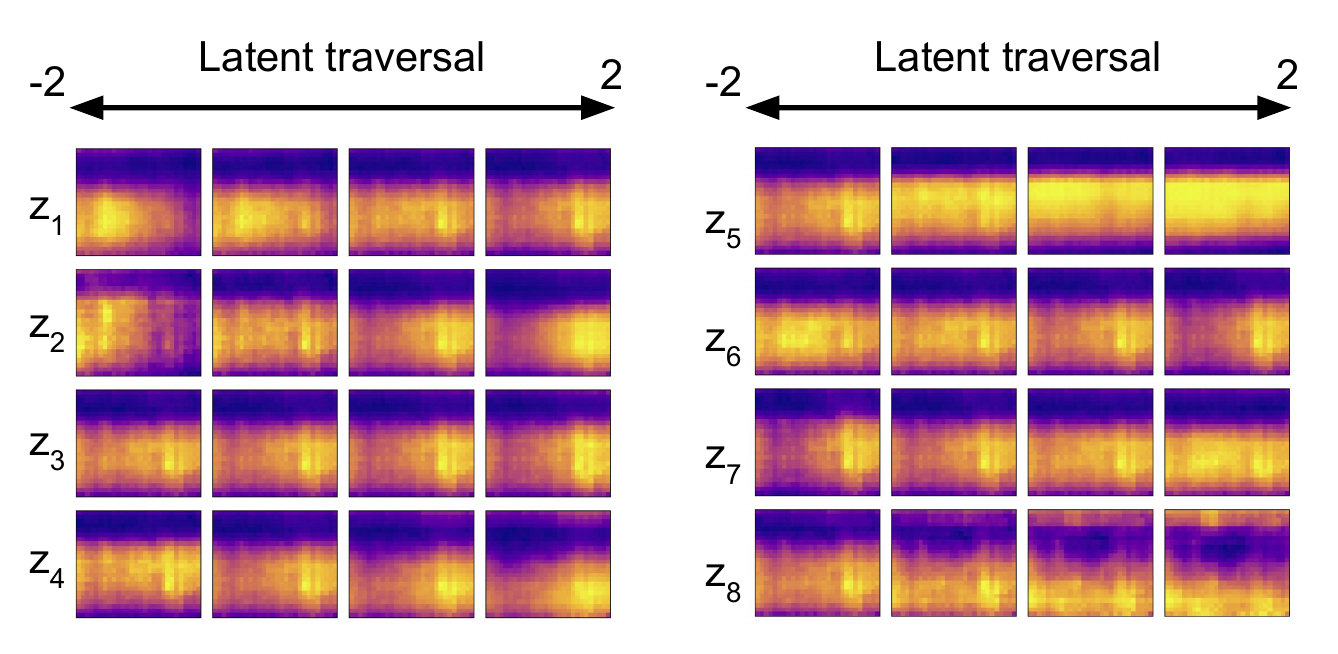}
    \caption{Latent traversals of pre-trained VAE model. Each row reconstructs an actigraphy map as the value of each latent dimension is traversed between [-2, 2] while keeping the values of all other latent variables fixed. Most of the variables show decreasing of daily energy, while $z_{8}$ shows the circadian phase change.}
    \label{traversals}
\end{figure}

In this work, we analyzed actigraphy data to estimate mental health outcomes after acute trauma. First, we used a convolutional VAE to extract unsupervised features from the four-week actigraphy data plots. The latent variables from the VAE model were fed into a logistic regression classifier for the binary classification task. We visualized what the different VAE latent representations learned by plotting their traversals as shown in Fig. \ref{traversals}. Then, we compared the performance with fully-supervised CNN-LSTM models and also used artificially generated data from the VAE model to enhance the performance of the CNN-LSTM approach.

We observed that when the unsupervised features extracted with the VAE model were combined with the physical health before the traumatic event (captured with the $SF-12$), the model achieved an AUC of $0.64$ and an accuracy of $0.60$ in differentiating healthy participants from participants showing PTSD and depression symptoms as determined by clinical surveys. When model was reduced to passive data only (by removing the $SF-12$ feature), the AUC dropped by  3\%, but the accuracy was unchanged. The model performance was also tested to identify participants with PTSD and sleep disturbance, as shown in Table \ref{VAEresults}. The CNN-LSTM model had an accuracy of $0.56$  and an AUC of $0.57$ in classifying healthy participants and participants showing PTSD and depression symptoms. By incorporating the artificial data generated by the VAE, the recall of the model increased from $0.45$ to $0.60$, while other metrics did not change.

\begin{table}[t!]
    \centering
    \caption{VAE model mental health outcome estimation performance. Results are reported as mean(standard deviation) of the external folds of each experiment.}
    \label{VAEresults}
    \begin{tabular}{@{}ccccccc@{}}
    \toprule
    \textbf{Outcome} & \textbf{Features} & \textbf{\begin{tabular}[c]{@{}c@{}}Num.\\ healthy/\\ unhealthy\end{tabular}} & \textbf{Acc.} & \textbf{AUC} & \textbf{Precision} & \textbf{Recall}\\ \midrule
    {\small PCL-5} & $z_{act}$ & 554/559 & 0.55(0.01) & 0.56(0.01) & 0.55(0.01) & 0.51(0.01)\\ \midrule
    \begin{tabular}[c]{@{}c@{}}{\small PanicSleep}\\ {\small PCL-5}\end{tabular} & $z_{act}$ & 520/149 & 0.59(0.02) & 0.61(0.03) & 0.59(0.02) & 0.57(0.02)\\ \midrule
    \begin{tabular}[c]{@{}c@{}}{\small PanicSleep}\\ {\small PCL-5}\\ {\small PROM-Dep8b}\end{tabular} & $z_{act}$ & 494/111 & 0.60(0.03) & 0.61(0.03) & 0.60(0.04) & 0.58(0.03)\\ \midrule
    \begin{tabular}[c]{@{}c@{}}{\small PanicSleep}\\ {\small PCL-5}\\ {\small PROM-Dep8b}\end{tabular} & \begin{tabular}[c]{@{}c@{}}$z_{act}$\\ {\small SF-12}\end{tabular} & 479/108 & 0.60(0.02) & 0.64(0.03) & 0.61(0.02) & 0.57(0.03)\\ \bottomrule
    \end{tabular}
\end{table}

\begin{table}[t!]
    \centering
    \caption{CNN-LSTM model mental health outcome estimation performance. Results are reported as mean(standard deviation) of the external folds of each experiment.}
    \label{CNNLSTMresults}
    \begin{tabular}{@{}ccccccc@{}}
    \toprule
    \textbf{Outcome} & \textbf{Features} & \textbf{\begin{tabular}[c]{@{}c@{}}Num.\\ healthy/\\ unhealthy\end{tabular}} & \textbf{Acc.} & \textbf{AUC} & \textbf{Precision} & \textbf{Recall} \\ \midrule
    {\small PCL-5} & $z_{act}$ & 554/559 & 0.52(0.01) & 0.53(0.01) & 0.53(0.02) & 0.40(0.04) \\ \midrule
    \begin{tabular}[c]{@{}c@{}}{\small PanicSleep}\\ {\small PCL-5}\end{tabular} & $z_{act}$ & 520/149 & 0.56(0.03) & 0.59(0.03) & 0.58(0.04) & 0.48(0.06)\\ \midrule
    \begin{tabular}[c]{@{}c@{}}{\small PanicSleep}\\ {\small PCL-5}\\ {\small PROM-Dep8b}\end{tabular} & $z_{act}$ & 494/111 & 0.56(0.04) & 0.57(0.04) & 0.58(0.05) & 0.45(0.08)\\ \bottomrule
    \end{tabular}
\end{table}

In conclusion, leveraging the unsupervised learning, the VAE model can extract more informative features and achieved higher accuracy compared to the CNN-LSTM. VAE approach compresses four-week worth of actigraphy data into an $8$ dimensional vector. Therefore, this method could result in immense memory savings for applications with more data streams or long-term studies and could be adapted to different and novel devices.

\begin{ack}
The funding for the study was provided by NIMH U01MH110925, the US Army Medical Research and Material Command, The One Mind Foundation, and The Mayday Fund. Verily Life Sciences and Mindstrong Health provided some of the hardware and software used to perform study assessments.
\end{ack}




\end{document}